\documentclass[preprint,12pt]{elsarticle}

\usepackage[margin=2.5cm]{geometry}
\usepackage{float}
\usepackage{amsmath,amssymb,amsfonts}
\usepackage{array}
\usepackage{graphicx}
\usepackage{booktabs}
\usepackage{threeparttable}
\usepackage{algorithm}
\usepackage{algpseudocode}
\usepackage{hyperref}
\usepackage[dvipsnames]{xcolor}
\usepackage{tikz}
\usetikzlibrary{arrows.meta,backgrounds,calc,fit,positioning}

\definecolor{satA}{RGB}{54,130,199}
\definecolor{satB}{RGB}{218,119,44}
\definecolor{satC}{RGB}{80,170,100}
\definecolor{satD}{RGB}{185,90,150}

\journal{Aerospace Science and Technology}

\begin{document}

\begin{frontmatter}

\title{Implicit Q-learning-bootstrapped ant colony optimization for maritime moving-target observation scheduling with agile satellites\tnoteref{tnote1}}

\tnotetext[tnote1]{%
\textcopyright{} 2026. This manuscript version is made available under the CC-BY-NC-ND 4.0 license
\url{https://creativecommons.org/licenses/by-nc-nd/4.0/}.
The Version of Record is available at
\url{https://doi.org/10.1016/j.ast.2026.113899}.}

\author[aff1]{He Wang}

\author[aff1]{Junyu Wu}

\author[aff2]{Yeye Liu}

\author[aff1]{Yifan Zhou}

\author[aff1]{Jie Zhang}

\author[aff1]{Hui Li}

\author[aff3]{Yanjie Song}

\author[aff1]{Liang Li\corref{cor1}}
\ead{liliang@hrbeu.edu.cn}
\cortext[cor1]{Corresponding author.}

\address[aff1]{College of Intelligent Science and Engineering, Harbin Engineering University, Harbin 150001, China}
\address[aff2]{School of Electrical and Control Engineering, North University of China, Taiyuan 030051, China}
\address[aff3]{School of Information Science and Technology, Dalian Maritime University, Dalian 116026, China}

\begin{abstract}
Maritime moving-target observation scheduling with agile Earth observation satellites is a dynamic, sequence-dependent combinatorial optimization problem. Sea-surface targets move continuously, causing feasible observation windows to vary with target motion and satellite orbital geometry. The scheduler must jointly determine task selection, satellite assignment, observation-window selection, and observation ordering under time-window, attitude-maneuvering, and onboard-resource constraints. This paper proposes an implicit Q-learning-bootstrapped ant colony optimization method, termed IQACO, for multi-satellite maritime moving-target observation scheduling. Rather than directly learning a task-selection policy, IQACO embeds an offline implicit Q-learning module into constructive ant colony optimization to adaptively adjust the pheromone factor, heuristic factor, and evaporation rate. A compact search-state representation captures pheromone distribution, current and historical-best solution quality, and iteration progress. During online scheduling, ant colony optimization constructs feasible observation sequences, while the learned policy adjusts the search behavior according to the current search state. Experiments on 14 scenarios with different scales and satellite configurations show that IQACO consistently outperforms the compared algorithms, improving the mean objective value over conventional ant colony optimization by 2.86\%--9.41\%. Further comparative and supplementary experiments demonstrate its effectiveness and robustness across different scheduling conditions and problem settings. These results indicate that offline value learning provides an effective adaptive search-control mechanism for constrained maritime moving-target observation scheduling.

\end{abstract}


\end{frontmatter}

\section{Introduction}

\begin{figure}[H]
	\centering
	\includegraphics[width=0.8\textwidth]{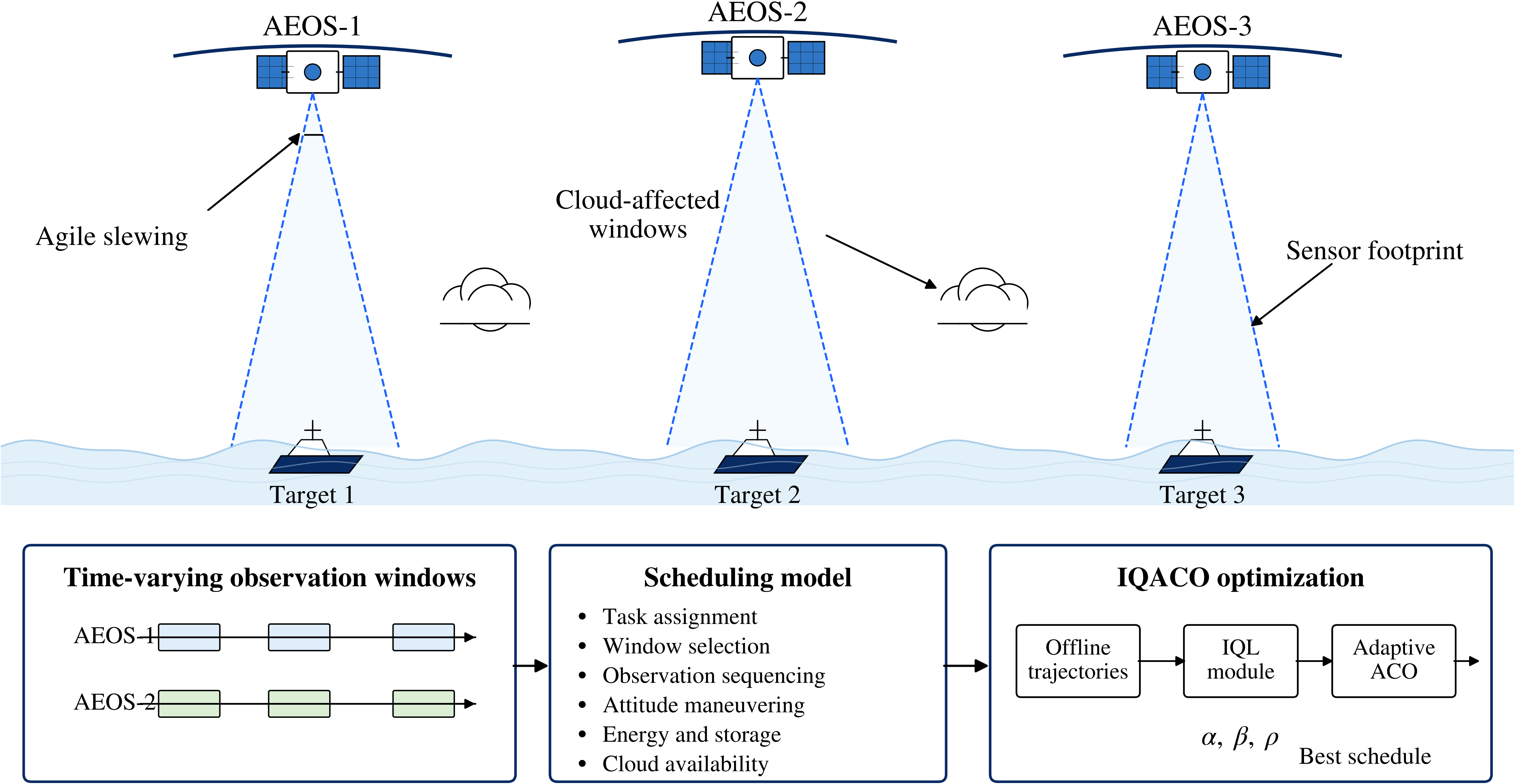}
\caption{Schematic of multi-satellite maritime moving-target scheduling, including time-varying observation windows, operational constraints, and the IQL-guided adaptive ACO loop.}
	\label{fig01}
\end{figure}

Maritime moving-target observation using agile Earth observation satellites (AEOSs) is important for vessel traffic monitoring, maritime search and rescue, illegal fishing detection, maritime security, and environmental surveillance. Compared with conventional satellites with limited pointing capability, AEOSs can rapidly slew their payloads toward sea-surface targets within limited visibility intervals. When multiple satellites are coordinated to observe numerous moving targets, the scheduling problem is no longer a simple target-selection problem, but a coupled decision-making problem involving satellite-task assignment, observation-window selection, observation sequencing, and feasibility checking under attitude-maneuvering and resource constraints~\cite{01,02,03}. The overall scheduling scenario is illustrated in Fig.~\ref{fig01}.

Maritime moving-target scheduling is more challenging than static-target scheduling. Sea-surface targets continuously change their relative geometry with satellite orbits and sensor footprints, making feasible observation windows strongly time dependent. In addition, schedule feasibility depends on the execution order of selected tasks because attitude-maneuvering time and maneuvering energy are required between consecutive observations. Frequent slews and high-resolution imaging further consume limited onboard energy and storage resources~\cite{05,06,07,08}. For optical payloads, ocean cloud cover may reduce the practical value of geometrically visible windows~\cite{09,10,11}. These factors jointly lead to a large-scale, sequence-dependent, and resource-constrained combinatorial optimization problem.

AEOS scheduling has been studied for more than two decades. Existing methods include exact optimization, heuristic and local-search methods, metaheuristics, and learning-based approaches~\cite{12}. Exact methods, such as integer programming, branch-and-bound, column generation, and constraint programming, can provide rigorous solutions for small or medium instances, but their computational cost increases rapidly when sequence-dependent transitions, onboard resources, and uncertainty are jointly considered~\cite{02,13,14}. Heuristic and local-search methods are efficient and easy to implement, but their performance often depends on hand-crafted rules and problem-specific assumptions~\cite{15,16}. Metaheuristics, such as GA, PSO, WOA, ACO, memetic algorithms, and hybrid neighborhood search, have therefore been widely used for large-scale AEOS scheduling~\cite{01,17,18,19,20}.

Among metaheuristics, ACO constructs solutions sequentially, with pheromone and heuristic information guiding each task-transition decision. Conventional ACO typically uses fixed parameters, including the pheromone factor $\alpha$, heuristic factor $\beta$, and evaporation rate $\rho$~\cite{21}. Fixed parameters may not maintain the same exploration--exploitation behavior when target density, satellite number, feasible-window distribution, and resource constraints vary across scenarios. This motivates adaptive parameter control based on the current search state.

Reinforcement learning (RL) has recently attracted attention in AEOS scheduling because it can learn scheduling policies or value functions from data~\cite{22,23,24,25,26,27}. However, many existing RL-based schedulers directly learn task-selection or sequence-generation policies. Such methods can involve large discrete action spaces and dynamically changing feasible action sets. Hard operational constraints also need to be incorporated into the decision process, for example through action masking, penalty terms, or repair procedures.

Online and offline RL differ mainly in how the training data are obtained. Online RL learns from interaction-generated experience, whereas offline RL learns from a fixed dataset collected beforehand. In the present framework, offline training separates policy learning from mission scheduling and avoids additional RL training during schedule execution. IQL is adopted because it can learn from a fixed dataset without explicitly maximizing over out-of-distribution actions~\cite{28,29,30}.

To address these issues, this article develops an implicit Q-learning-bootstrapped ant colony optimization method, termed IQACO, for multi-satellite maritime moving-target observation scheduling. Unlike RL schedulers that directly generate task or sequence decisions, IQACO uses the learned policy to adjust the ACO parameters $\alpha$, $\beta$, and $\rho$. The policy therefore has a three-dimensional action space, while feasible schedule construction remains the responsibility of the ACO decoder. Time-window, attitude-maneuvering, energy, and storage constraints are checked during schedule construction. In this way, IQL is used for search control rather than for direct schedule generation.

The main contributions of this article are summarized as follows:
\begin{enumerate}
	\item A multi-satellite maritime moving-target observation scheduling model is formulated by integrating time-varying observation windows, satellite-task-window assignment, sequence-dependent attitude maneuvering, onboard energy and storage resources, and cloud-affected availability.
	\item An IQACO method is developed to adjust ACO parameters using offline IQL. The learned policy updates the pheromone factor, heuristic factor, and evaporation rate according to the current search state, thereby adapting the exploration--exploitation behavior during feasible schedule construction.
	\item Experiments on 14 scenarios with different problem scales and satellite configurations evaluate convergence behavior, final weighted objective values, objective-weight sensitivity, and IQL training behavior.
\end{enumerate}

The remainder of this article is organized as follows. Section~\ref{sec02} formulates the scheduling model. Section~\ref{sec03} presents the proposed IQACO method. Section~\ref{sec04} reports the simulation experiments and result analysis. Section~\ref{sec05} concludes this article.

\section{Scheduling Model}\label{sec02}
\subsection{Problem Description}

We consider multi-satellite maritime moving-target observation scheduling over a finite planning horizon $T_H$. The scheduling objects include a set of agile satellites and a set of maritime moving targets. Before optimization, target trajectories and satellite ephemerides are propagated to generate feasible task--satellite observation windows. The scheduler then determines task selection, satellite assignment, observation-window selection, and task ordering on each satellite under time-window, attitude-maneuvering, onboard energy, storage-capacity, and cloud-affected availability constraints.

Compared with static-target scheduling, maritime moving-target observation has stronger temporal and spatial variability. The relative geometry between satellites and targets changes with both vessel motion and orbital motion, making feasible observation windows time dependent. A geometrically visible window may also have low practical availability due to ocean cloud cover. For agile satellites, schedule feasibility further depends on the execution order of selected tasks because attitude-maneuvering time and energy are required between consecutive observations. Therefore, the problem is a sequence-dependent and resource-constrained combinatorial optimization problem.

\subsection{Sets and Parameters}

The main notation used in the scheduling model is listed in Table~\ref{tab01}. An original moving target may have multiple observation requirements within the planning horizon; these requirements are expanded into independent scheduling tasks. If task $i$ cannot be observed by satellite $s$, the corresponding feasible-window set $\mathcal{W}_{i,s}$ is empty.

Two binary variables are used to describe the scheduling decision. The variable $x_{i,s,k}$ indicates whether task $i$ is assigned to satellite $s$ and executed in window $k$. It is set to one if the corresponding task--satellite--window combination is selected, and zero otherwise: 
\begin{equation} 
	x_{i,s,k} = \begin{cases} 1, & \text{selected},\\ 0, & \text{otherwise}. \end{cases}
\end{equation} 

The variable $y_{ij}^{s}$ describes the immediate-successor relation in the task sequence of satellite $s$. It is set to one if satellite $s$ executes task $j$ immediately after task $i$, and zero otherwise: 
\begin{equation} 
	y_{ij}^{s} = \begin{cases} 1, & \text{if } j \text{ immediately follows } i,\\ 0, & \text{otherwise}. \end{cases} 
\end{equation} 

Thus, $x_{i,s,k}$ determines task selection, satellite assignment, and observation-window selection, while $y_{ij}^{s}$ describes the task order on each satellite. The latter is also used to compute the attitude-maneuvering time and energy between consecutive observations.

\begin{table}[!h]
	\centering
	\caption{Notation used in the scheduling model}
	\label{tab01}
	\footnotesize
	\setlength{\tabcolsep}{3pt}
	\renewcommand{\arraystretch}{1.08}
	\begin{tabular}{p{0.26\columnwidth}p{0.62\columnwidth}}
		\toprule
		Symbol & Description \\
		\midrule
		$\mathcal{T}$, $\mathcal{S}$ & Sets of scheduling tasks and agile satellites \\
		$\mathcal{W}_{i,s}$ & Feasible observation-window set for task $i$ and satellite $s$ \\
		$i,j$ & Task indices \\
		$s$ & Satellite index \\
		$k,l$ & Observation-window indices \\
		$N_T$, $N_S$ & Numbers of tasks and satellites \\
		$T_H$ & Planning-horizon length \\
		$t_{i,s,k}^{\mathrm{start}}$ & Start time of window $k$ \\
		$t_{i,s,k}^{\mathrm{end}}$ & End time of window $k$ \\
		$t_{i,s,k}^{\mathrm{obs}}$ & Observation start time \\
		$d_i$, $p_i$ & Observation duration and nominal benefit of task $i$ \\
		$c_{i,s,k}$, $c_{\min}$ & Cloud-affected availability factor and threshold \\
		$\theta_{ij}^{s}$ & Slewing angle from task $i$ to task $j$ \\
		$T_{ij}^{s,\mathrm{man}}$ & Maneuvering time from task $i$ to task $j$ \\
		$\omega_s^{\max}$ & Maximum angular rate of satellite $s$ \\
		$a_s^{\max}$ & Maximum angular acceleration of satellite $s$ \\
		$e_{i,s,k}^{\mathrm{obs}}$ & Imaging energy \\
		$e_{ij}^{s,\mathrm{man}}$ & Maneuvering energy \\
		$E_s^{\max}$, $E_s^{\mathrm{use}}$ & Energy capacity and energy use of satellite $s$ \\
		$m_{i,s,k}$, $M_s^{\max}$ & Generated data volume and storage capacity \\
		$x_{i,s,k}$ & Task-window assignment variable \\
		$y_{ij}^{s}$ & Immediate-successor variable \\
		$F$, $F_p$, $F_e$, $F_b$ & Objective value and its observation-benefit, energy-efficiency, and workload-balance terms \\
		$\eta_1$, $\eta_2$, $\eta_3$ & Objective weights of the three normalized terms \\
		$L_s$, $\bar{L}$ & Workload of satellite $s$ and average workload \\
		\bottomrule
	\end{tabular}
\end{table}

\subsection{Constraints}

A feasible schedule must satisfy operational constraints related to task assignment, observation windows, task sequencing, attitude maneuvering, onboard resources, and cloud-affected availability. These constraints are used as feasibility rules in the constructive decoder. 
\subsubsection{Task Uniqueness Constraint}

Each task is executed at most once:
\begin{equation}
	\sum_{s\in\mathcal{S}}\sum_{k\in\mathcal{W}_{i,s}} x_{i,s,k} \leq 1,
	\quad \forall i\in\mathcal{T}.
\end{equation}

\subsubsection{Time-Window Feasibility Constraint}

The observation interval must be contained within the selected window:
\begin{subequations}
	\begin{align}
		t_{i,s,k}^{\mathrm{start}} &\leq t_{i,s,k}^{\mathrm{obs}}, \\
		t_{i,s,k}^{\mathrm{obs}}+d_i &\leq t_{i,s,k}^{\mathrm{end}} .
	\end{align}
\end{subequations}

These inequalities are enforced only for selected task--satellite--window nodes with $x_{i,s,k}=1$.

\subsubsection{Sequence-Linking Constraint}

The sequence variable $y_{ij}^{s}$ is valid only when both tasks $i$ and $j$ are assigned to satellite $s$:
\begin{align}
	y_{ij}^{s} &\leq \sum_{k\in\mathcal{W}_{i,s}} x_{i,s,k},
	&& \forall i,j\in\mathcal{T},\ i\neq j,\ s\in\mathcal{S}, \\
	y_{ij}^{s} &\leq \sum_{l\in\mathcal{W}_{j,s}} x_{j,s,l},
	&& \forall i,j\in\mathcal{T},\ i\neq j,\ s\in\mathcal{S}.
\end{align}

Each scheduled task has at most one immediate successor and predecessor:
\begin{align}
	\sum_{\substack{j\in\mathcal{T}\\j\neq i}} y_{ij}^{s}
	&\leq \sum_{k\in\mathcal{W}_{i,s}} x_{i,s,k},
	&& \forall i\in\mathcal{T},\ s\in\mathcal{S}, \\
	\sum_{\substack{i\in\mathcal{T}\\i\neq j}} y_{ij}^{s}
	&\leq \sum_{l\in\mathcal{W}_{j,s}} x_{j,s,l},
	&& \forall j\in\mathcal{T},\ s\in\mathcal{S}.
\end{align}

These inequalities link task-assignment variables with immediate-successor relations. Complete satellite-specific sequences are generated by the constructive ACO procedure, which checks successor relations, temporal feasibility, and resource feasibility during schedule construction.

\subsubsection{Attitude Maneuvering Constraint}\label{subsec:maneuver}

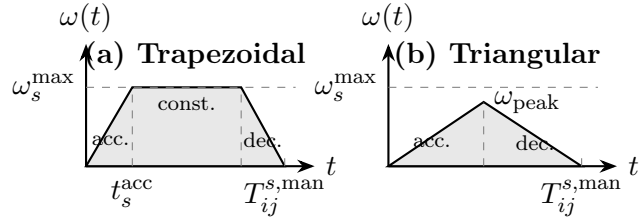
\begin{figure}[H]
	\centering
	\begin{tikzpicture}[
		x=0.72cm,
		y=0.72cm,
		>=Stealth,
		font=\small,
		axis/.style={->, thick},
		prof/.style={thick},
		guide/.style={dashed, thin, gray},
		fillprof/.style={fill=gray!18, draw=black, thick}
		]

		\begin{scope}[xshift=0cm]
			
			\draw[axis] (0,0) -- (4.2,0) node[right] {$t$};
			\draw[axis] (0,0) -- (0,2.25) node[above] {$\omega(t)$};

			\def\tacc{0.85}
			\def\tflatend{2.85}
			\def\tend{3.65}
			\def\wmax{1.45}

			\filldraw[fillprof] (0,0) -- (\tacc,\wmax) -- (\tflatend,\wmax) -- (\tend,0) -- cycle;

			\draw[guide] (0,\wmax) -- (3.85,\wmax);
			\draw[guide] (\tacc,0) -- (\tacc,\wmax);
			\draw[guide] (\tflatend,0) -- (\tflatend,\wmax);
			\draw[guide] (\tend,0) -- (\tend,0.16);

			\node[left] at (0,\wmax) {$\omega_s^{\max}$};
			\node[below=1pt] at (\tacc,0) {$t_s^{\mathrm{acc}}$};
			\node[below=1pt] at (\tend,0) {$T_{ij}^{s,\mathrm{man}}$};
			\node[font=\small\bfseries] at (2.0,2.03) {(a) Trapezoidal};

			\node at (0.45,0.48) {\scriptsize acc.};
			\node at (1.85,1.15) {\scriptsize const.};
			\node at (3.25,0.48) {\scriptsize dec.};
		\end{scope}

		\begin{scope}[xshift=4cm]
			\draw[axis] (0,0) -- (4.2,0) node[right] {$t$};
			\draw[axis] (0,0) -- (0,2.25) node[above] {$\omega(t)$};

			\def\tmid{1.75}
			\def\tendb{3.55}
			\def\wpeak{1.18}
			\def\wmaxb{1.45}

			\filldraw[fillprof] (0,0) -- (\tmid,\wpeak) -- (\tendb,0) -- cycle;

			\draw[guide] (0,\wmaxb) -- (3.85,\wmaxb);
			\draw[guide] (\tmid,0) -- (\tmid,\wpeak);
			\draw[guide] (\tendb,0) -- (\tendb,0.16);

			\node[left] at (0,\wmaxb) {$\omega_s^{\max}$};
			\node[below=1pt] at (\tendb,0) {$T_{ij}^{s,\mathrm{man}}$};
			\node[right] at (\tmid,\wpeak) {$\omega_{\mathrm{peak}}$};
			\node[font=\small\bfseries] at (2.0,2.03) {(b) Triangular};

			\node at (0.80,0.43) {\scriptsize acc.};
			\node at (2.70,0.43) {\scriptsize dec.};
		\end{scope}
		
	\end{tikzpicture}
	\vspace{-0.8em}
	\caption{Attitude-maneuvering angular-rate profiles of a rate- and acceleration-limited satellite: (a) trapezoidal profile, used when the slewing angle is large enough to reach the maximum angular rate; (b) triangular profile, used otherwise.}
	\label{fig_maneuver_profile}
\end{figure}

For agile satellites, an attitude maneuver is required between two consecutive observations assigned to the same satellite. The required slewing angle $\theta_{ij}^{s}$ is computed from the angular separation between the payload pointing directions of tasks $i$ and $j$. A rate- and acceleration-limited maneuvering model is used, as illustrated in Fig.~\ref{fig_maneuver_profile}. If the required angle is large enough, the satellite reaches the maximum angular rate and follows a trapezoidal profile; otherwise, it follows a triangular profile.

Let $\omega_s^{\max}$ and $a_s^{\max}$ denote the maximum angular rate and maximum angular acceleration of satellite $s$, respectively. The acceleration or deceleration time is
\begin{equation}
	t_s^{\mathrm{acc}}=\frac{\omega_s^{\max}}{a_s^{\max}} .
\end{equation}

When the satellite can reach the maximum angular rate, the maneuvering time is
\begin{equation}
	T_{ij}^{s,\mathrm{man}}
	=
	2t_s^{\mathrm{acc}}
	+
	\frac{\theta_{ij}^{s}-(\omega_s^{\max})^2/a_s^{\max}}
	{\omega_s^{\max}},
\end{equation}
where $\theta_{ij}^{s}\geq(\omega_s^{\max})^2/a_s^{\max}$. When the maximum angular rate cannot be reached, the maneuvering time is
\begin{equation}
	T_{ij}^{s,\mathrm{man}}
	=
	2\sqrt{\frac{\theta_{ij}^{s}}{a_s^{\max}}},
\end{equation}
where $\theta_{ij}^{s}<(\omega_s^{\max})^2/a_s^{\max}$.

When task $j$ immediately follows task $i$ on satellite $s$ and their selected windows are $k$ and $l$, respectively, the observation start times must satisfy
\begin{equation}
	t_{j,s,l}^{\mathrm{obs}}
	\geq
	t_{i,s,k}^{\mathrm{obs}}+d_i+T_{ij}^{s,\mathrm{man}}.
\end{equation}
This constraint reserves sufficient transition time between consecutive observations on the same satellite.

\subsubsection{Onboard Energy Constraint}

The total energy use, including imaging and maneuvering energy, must not exceed the onboard capacity:
\begin{equation}
	\begin{aligned}
		&\sum_{i\in\mathcal{T}}\sum_{k\in\mathcal{W}_{i,s}}
		e_{i,s,k}^{\mathrm{obs}}x_{i,s,k}
		+
		\sum_{\substack{i,j\in\mathcal{T}\\i\neq j}}
		e_{ij}^{s,\mathrm{man}}y_{ij}^{s}
		\leq E_s^{\max},
		\quad \forall s\in\mathcal{S}.
	\end{aligned}
\end{equation}

\subsubsection{Onboard Storage Constraint}

The generated data volume must not exceed the storage capacity:
\begin{equation}
	\sum_{i\in\mathcal{T}}\sum_{k\in\mathcal{W}_{i,s}}
	m_{i,s,k}x_{i,s,k}
	\leq M_s^{\max},
	\quad \forall s\in\mathcal{S}.
\end{equation}

\subsubsection{Cloud-Availability Constraint}

For optical observations, a geometrically visible window may still have low practical value due to cloud cover. Let $c_{i,s,k}$ denote the cloud-affected availability of executing task $i$ by satellite $s$ in window $k$. Candidate windows with availability lower than the minimum acceptable threshold $c_{\min}$ are excluded:
\begin{equation}
	x_{i,s,k}=0,
	\quad \text{if} \quad c_{i,s,k}<c_{\min}.
\end{equation}

For retained windows, $c_{i,s,k}$ is further used as a benefit attenuation factor in the objective function.

\subsection{Objective Function}

Under the above feasibility constraints, the objective is to maximize the overall observation performance of the satellite constellation. Three normalized components are considered: observation benefit, energy efficiency, and workload balance. The objective function is formulated as
\begin{equation}
    \max F=\eta_1F_p+\eta_2F_e+\eta_3F_b,
\end{equation}
where $F_p$, $F_e$, and $F_b$ denote the observation-benefit, energy-efficiency, and workload-balance terms, respectively. The objective weights satisfy
\begin{equation}
    \eta_1+\eta_2+\eta_3=1,\qquad
    \eta_1,\eta_2,\eta_3\geq0.
\end{equation}

The observation-benefit term is defined as
\begin{equation}
    F_p=
    \frac{
        \sum_{i\in\mathcal{T}}
        \sum_{s\in\mathcal{S}}
        \sum_{k\in\mathcal{W}_{i,s}}
        p_i c_{i,s,k}x_{i,s,k}
    }{
        \sum_{i\in\mathcal{T}}p_i
    }.
\end{equation}
The numerator is the effective benefit collected by the schedule, while the denominator is the nominal total benefit of all tasks. Hence, $F_p\in[0,1]$, with a larger value indicating a larger fraction of the nominal benefit obtained.

For satellite $s$, the total energy use consists of imaging and attitude-maneuvering energy:
\begin{equation}
    E_s^{\mathrm{use}}
    =
    \sum_{i\in\mathcal{T}}
    \sum_{k\in\mathcal{W}_{i,s}}
    e_{i,s,k}^{\mathrm{obs}}x_{i,s,k}
    +
    \sum_{\substack{i,j\in\mathcal{T}\\i\neq j}}
    e_{ij}^{s,\mathrm{man}}y_{ij}^{s}.
\end{equation}
The energy-efficiency term is
\begin{equation}
    F_e
    =
    1-
    \frac{
        \sum_{s\in\mathcal{S}}E_s^{\mathrm{use}}
    }{
        \sum_{s\in\mathcal{S}}E_s^{\max}
    }.
\end{equation}
The total energy consumption is normalized by the total available onboard energy. Under the energy-capacity constraint, $F_e\in[0,1]$, and a larger value corresponds to lower relative energy consumption.

The workload of satellite $s$ is
\begin{equation}
    L_s=
    \sum_{i\in\mathcal{T}}
    \sum_{k\in\mathcal{W}_{i,s}}
    d_i x_{i,s,k},
\end{equation}
and the average workload is
\begin{equation}
    \bar{L}
    =
    \frac{1}{N_S}
    \sum_{s\in\mathcal{S}}L_s.
\end{equation}
The workload-balance term is defined as
\begin{equation}
    F_b
    =
    \left(
    1+
    \frac{
        \sqrt{
            \frac{1}{N_S}
            \sum_{s\in\mathcal{S}}
            (L_s-\bar{L})^2
        }
    }{
        \bar{L}+\epsilon
    }
    \right)^{-1},
\end{equation}
where $\epsilon$ is a small positive constant used to avoid division by zero. The ratio inside the parentheses measures workload dispersion relative to the average workload. Thus, $F_b=1$ for a perfectly balanced workload distribution and decreases as the workload becomes more uneven.

All three terms are dimensionless. The weights $(\eta_1,\eta_2,\eta_3)$ therefore represent the relative preference among observation benefit, energy efficiency, and workload balance.

\section{Method}\label{sec03}

\subsection{Method Overview}

\begin{figure}[H]
	\centering
	\resizebox{0.99\textwidth}{!}{%
	\begin{tikzpicture}[
		font=\small,
		node distance=0.58cm and 0.62cm,
		>={Stealth[length=2.0mm,width=1.2mm]},
		block/.style={
			draw,
			rounded corners=2pt,
			align=center,
			minimum width=2.10cm,
			minimum height=0.84cm,
			inner sep=3pt,
			line width=0.45pt
		},
		normal/.style={
			block,
			fill=white,
			draw=black!70
		},
		data/.style={
			block,
			fill=violet!8,
			draw=violet!45!black
		},
		train/.style={
			block,
			fill=orange!10,
			draw=orange!50!black
		},
		policybox/.style={
			block,
			fill=yellow!10,
			draw=yellow!45!black
		},
		onlinekey/.style={
			block,
			fill=green!10,
			draw=green!40!black
		},
		groupbox/.style={
			draw,
			rounded corners=3pt,
			inner sep=0.18cm,
			line width=0.55pt
		},
		arrow/.style={
			-{Stealth[length=2.0mm,width=1.2mm]},
			line width=0.55pt,
			draw=black!75
		},
		feedback/.style={
			-{Stealth[length=2.0mm,width=1.2mm]},
			dashed,
			line width=0.55pt,
			draw=black!60
		},
		policyarrow/.style={
			-{Stealth[length=2.0mm,width=1.2mm]},
			line width=0.55pt,
			draw=black!75
		}
		]

		\node[normal] (scenarios) {Training\\scenarios};
		\node[normal, right=0.62cm of scenarios] (aco_runs) {Exploratory\\ACO runs};
		\node[data, right=0.62cm of aco_runs] (dataset) {Offline dataset\\$(\mathbf{s}_t,\mathbf{a}_t,R_t,\mathbf{s}_{t+1},d_t)$};
		\node[train, right=0.62cm of dataset] (iql_train) {IQL training\\$Q,V,\pi$ networks};
		\node[policybox, right=0.62cm of iql_train] (policy) {Trained policy\\$\pi_{\phi}(\mathbf{s})$};
		
		\draw[arrow] (scenarios) -- (aco_runs);
		\draw[arrow] (aco_runs) -- (dataset);
		\draw[arrow] (dataset) -- (iql_train);
		\draw[arrow] (iql_train) -- (policy);

		\node[normal, below=1.20cm of aco_runs] (test) {Test\\scenario};
		\node[normal, right=0.62cm of test] (state) {Extract\\search state\\$\mathbf{s}_t$};
		\node[train, right=0.62cm of state] (action) {Policy inference\\$\mathbf{a}_t=\pi_{\phi}(\mathbf{s}_t)$};
		\node[normal, right=0.62cm of action] (params) {Update\\$\alpha,\beta,\rho$};
		\node[onlinekey, right=0.62cm of params] (construct) {ACO schedule\\construction};
		
		\node[normal, below=0.76cm of construct] (update) {Evaluate schedules\\and update pheromone};
		\node[onlinekey, left=0.62cm of update] (best) {Best schedule\\$\Pi^{\ast}$};
		
		\draw[arrow] (test) -- (state);
		\draw[arrow] (state) -- (action);
		\draw[arrow] (action) -- (params);
		\draw[arrow] (params) -- (construct);
		\draw[arrow] (construct) -- (update);
		\draw[arrow] (update) -- (best);

		\draw[feedback]
		(update.west) to[out=200,in=-20,looseness=1.15] (state.south);
		
		\draw[policyarrow] (policy.south) -- ++(0,-0.36cm) -| (action.north);

		\begin{scope}[on background layer]
			\node[groupbox,
			draw=blue!35!black,
			fill=blue!2,
			fit=(scenarios)(aco_runs)(dataset)(iql_train)(policy),
			label={[font=\small\bfseries,text=blue!40!black]above:Offline IQL training stage}
			] (offlinebox) {};
			
			\node[groupbox,
			draw=green!30!black,
			fill=green!2,
			fit=(test)(state)(action)(params)(construct)(update)(best),
			label={[font=\small\bfseries,text=green!35!black]above:Online IQACO scheduling stage}
			] (onlinebox) {};
		\end{scope}
		
	\end{tikzpicture}
	}
	\vspace{-3pt}
\caption{Workflow of the proposed IQACO method. The offline stage collects transitions with exploratory parameter adjustments and trains the IQL networks; the online stage uses the trained policy to adjust the pheromone factor, heuristic factor, and evaporation rate for feasible schedule construction.}
	\label{fig030}
\end{figure}
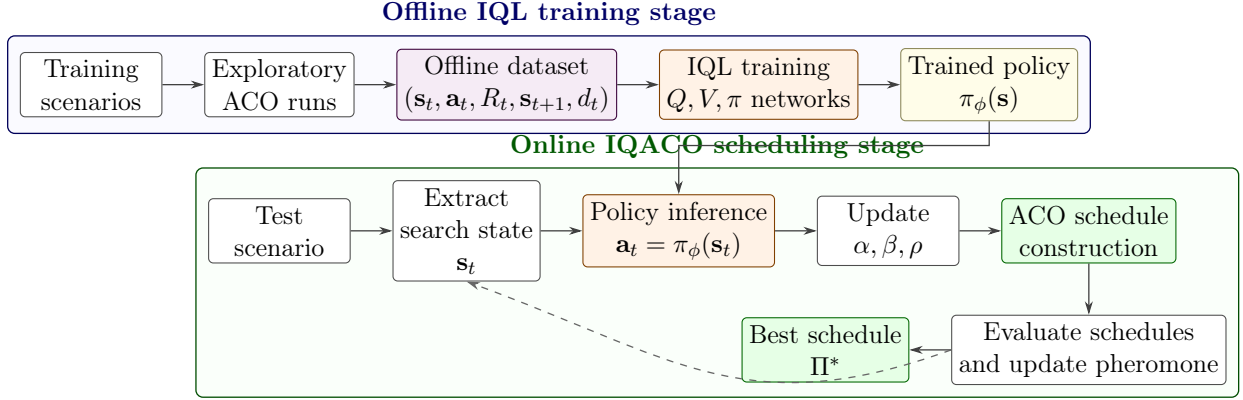

IQACO consists of an offline IQL training stage and an online ACO scheduling stage, as shown in Fig.~\ref{fig030}. In the offline stage, ACO is executed on training scenarios with exploratory parameter adjustments, and the resulting transitions $(\mathbf{s}_t,\mathbf{a}_t,R_t,\mathbf{s}_{t+1},d_t)$ are collected to train an IQL policy. In the online stage, ACO constructs feasible schedules at each iteration, updates the best solution and pheromone matrix, and then uses the trained policy to adjust $\alpha$, $\beta$, and $\rho$ according to the current search state.

Unlike direct RL schedulers that output task-selection or sequence decisions, IQACO learns only the continuous parameter adjustments used by ACO. Feasible schedules are constructed by the constraint-aware ACO procedure, so the learned policy does not generate task sequences directly. IQL therefore serves as a search controller rather than as a stand-alone scheduling policy.

\subsection{Solution Encoding and ACO-Based Schedule Construction}

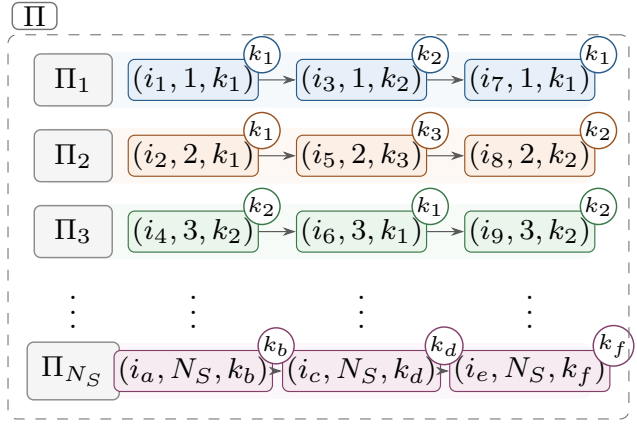
\begin{figure}[H]
	\centering
	\resizebox{0.5\columnwidth}{!}{%
		\begin{tikzpicture}[
			font=\scriptsize,
			>=Stealth,
			task/.style={
				draw=#1!70!black,
				rounded corners=1.8pt,
				minimum width=1.15cm,
				minimum height=0.40cm,
				align=center,
				inner sep=1.0pt,
				line width=0.32pt,
				fill=#1!12
			},
			task/.default=gray,
			badge/.style={
				circle,
				draw=#1!70!black,
				fill=white,
				inner sep=0.20pt,
				minimum size=0.22cm,
				font=\tiny\bfseries
			},
			rowlabel/.style={
				draw=black!45,
				rounded corners=1.8pt,
				fill=black!4,
				minimum width=0.76cm,
				minimum height=0.38cm,
				font=\scriptsize\bfseries,
				align=center
			},
			arrow/.style={
				-{Stealth[length=1.30mm,width=0.85mm]},
				line width=0.32pt,
				draw=black!65
			},
			dotnode/.style={
				font=\scriptsize,
				align=center
			},
			rowband/.style={
				draw=none,
				fill=#1!5,
				rounded corners=2pt
			},
			globaltag/.style={
				draw=black!55,
				fill=white,
				rounded corners=1.8pt,
				inner xsep=3.0pt,
				inner ysep=1.0pt,
				font=\scriptsize\bfseries
			}
			]

			\def\xlabel{0.35}
			\def\xone{1.52}
			\def\xtwo{3.16}
			\def\xthree{4.80}
			
			\def\yone{0.00}
			\def\ytwo{-0.74}
			\def\ythree{-1.48}
			\def\ydots{-2.16}
			\def\yns{-2.84}

			\node[rowband=satA, minimum width=5.35cm, minimum height=0.54cm] at (2.72,\yone) {};
			\node[rowband=satB, minimum width=5.35cm, minimum height=0.54cm] at (2.72,\ytwo) {};
			\node[rowband=satC, minimum width=5.35cm, minimum height=0.54cm] at (2.72,\ythree) {};
			\node[rowband=satD, minimum width=5.35cm, minimum height=0.54cm] at (2.72,\yns) {};

			\node[rowlabel] (p1) at (\xlabel,\yone) {$\Pi_1$};
			\node[rowlabel] (p2) at (\xlabel,\ytwo) {$\Pi_2$};
			\node[rowlabel] (p3) at (\xlabel,\ythree) {$\Pi_3$};
			\node[dotnode]  (pd) at (\xlabel,\ydots) {$\vdots$};
			\node[rowlabel] (pn) at (\xlabel,\yns) {$\Pi_{N_S}$};

			\node[task=satA] (a11) at (\xone,\yone) {$(i_1,1,k_1)$};
			\node[task=satA] (a12) at (\xtwo,\yone) {$(i_3,1,k_2)$};
			\node[task=satA] (a13) at (\xthree,\yone) {$(i_7,1,k_1)$};
			\draw[arrow] (a11) -- (a12);
			\draw[arrow] (a12) -- (a13);
			\node[badge=satA] at ($(a11.north east)+(0.03,0.04)$) {$k_1$};
			\node[badge=satA] at ($(a12.north east)+(0.03,0.04)$) {$k_2$};
			\node[badge=satA] at ($(a13.north east)+(0.03,0.04)$) {$k_1$};

			\node[task=satB] (a21) at (\xone,\ytwo) {$(i_2,2,k_1)$};
			\node[task=satB] (a22) at (\xtwo,\ytwo) {$(i_5,2,k_3)$};
			\node[task=satB] (a23) at (\xthree,\ytwo) {$(i_8,2,k_2)$};
			\draw[arrow] (a21) -- (a22);
			\draw[arrow] (a22) -- (a23);
			\node[badge=satB] at ($(a21.north east)+(0.03,0.04)$) {$k_1$};
			\node[badge=satB] at ($(a22.north east)+(0.03,0.04)$) {$k_3$};
			\node[badge=satB] at ($(a23.north east)+(0.03,0.04)$) {$k_2$};

			\node[task=satC] (a31) at (\xone,\ythree) {$(i_4,3,k_2)$};
			\node[task=satC] (a32) at (\xtwo,\ythree) {$(i_6,3,k_1)$};
			\node[task=satC] (a33) at (\xthree,\ythree) {$(i_9,3,k_2)$};
			\draw[arrow] (a31) -- (a32);
			\draw[arrow] (a32) -- (a33);
			\node[badge=satC] at ($(a31.north east)+(0.03,0.04)$) {$k_2$};
			\node[badge=satC] at ($(a32.north east)+(0.03,0.04)$) {$k_1$};
			\node[badge=satC] at ($(a33.north east)+(0.03,0.04)$) {$k_2$};

			\node[dotnode] (d1) at (\xone,\ydots) {$\vdots$};
			\node[dotnode] (d2) at (\xtwo,\ydots) {$\vdots$};
			\node[dotnode] (d3) at (\xthree,\ydots) {$\vdots$};

			\node[task=satD] (an1) at (\xone,\yns) {$(i_a,N_S,k_b)$};
			\node[task=satD] (an2) at (\xtwo,\yns) {$(i_c,N_S,k_d)$};
			\node[task=satD] (an3) at (\xthree,\yns) {$(i_e,N_S,k_f)$};
			\draw[arrow] (an1) -- (an2);
			\draw[arrow] (an2) -- (an3);
			\node[badge=satD] at ($(an1.north east)+(0.03,0.04)$) {$k_b$};
			\node[badge=satD] at ($(an2.north east)+(0.03,0.04)$) {$k_d$};
			\node[badge=satD] at ($(an3.north east)+(0.03,0.04)$) {$k_f$};

			\node[
			draw=black!55,
			dashed,
			rounded corners=2.4pt,
			line width=0.30pt,
			fit=(p1)(a13)(pn)(an3),
			inner sep=0.18cm
			] (box) {};

			\node[
			globaltag,
			anchor=south west
			] at ([xshift=0.04cm,yshift=0.04cm]box.north west)
			{$\Pi$};
			
		\end{tikzpicture}%
}
\caption{Hierarchical solution encoding. The global schedule is decomposed into chronological satellite-specific sequences of task--satellite--window nodes; the badge of each node marks the selected observation window.}
	\label{fig040}
\end{figure}

A candidate observation is represented by a task--satellite--window node $z=(i,s,k)$, where $i$, $s$, and $k$ denote the task, satellite, and feasible observation window, respectively. Candidate nodes are pre-filtered according to visibility, time-window feasibility, and cloud-affected availability. A complete schedule is represented by the collection of ordered satellite-specific node sequences $\Pi=\{\Pi_1,\Pi_2,\ldots,\Pi_{N_S}\}$, as shown in Fig.~\ref{fig040}. Each $\Pi_s$ preserves the chronological observation sequence assigned to satellite $s$. Because each node records the selected task, satellite, and observation window, predecessor--successor relations can be recovered directly for maneuver-time, maneuver-energy, temporal-feasibility, and resource-feasibility checks. This representation captures task selection, satellite assignment, window selection, and observation ordering within the constructive search.

Each ant incrementally selects schedulable tasks from a feasible candidate set $\mathcal{C}^m$ that excludes candidates violating operational constraints. Algorithm~\ref{alg01} summarizes this construction. The evaporation rate $\rho$ is applied only during global pheromone updating, not during single-ant construction.

For the $m$th ant, the next candidate node $z=(i_z,s_z,k_z)\in\mathcal{C}^m$ is selected with probability
\begin{equation}
	P_m(z|u)
	=
	\frac{
		[\tau_{u,i_z}]^{\alpha}
		[\eta_m(u,z)]^{\beta}
	}{
		\sum_{q\in\mathcal{C}^m}
		[\tau_{u_q,i_q}]^{\alpha}
		[\eta_m(u_q,q)]^{\beta}
	},
\end{equation}
where $u$ is the latest scheduled task on satellite $s_z$, $\tau_{u,i_z}$ is the task-level pheromone intensity from $u$ to candidate task $i_z$, and $\eta_m(u,z)$ is the heuristic value of node $z$ under the current partial schedule. For another candidate node $q=(i_q,s_q,k_q)$ in the denominator, $u_q$ denotes the latest scheduled task on satellite $s_q$. The parameters $\alpha$ and $\beta$ control the relative influence of pheromone information and heuristic information.

\begin{algorithm}[!t]
	\caption{Schedule Construction by One Ant}
	\label{alg01}
	\begin{algorithmic}[1]
		\Require Candidate observation nodes, pheromone matrix $\boldsymbol{\tau}$, and ACO parameters $\alpha$ and $\beta$
		\Ensure A feasible schedule $\Pi^m$
		
		\State Initialize $\Pi^m\leftarrow\emptyset$, $\mathcal{U}\leftarrow\mathcal{T}$, and satellite states with virtual initial nodes
		\While{true}
		\State Build the feasible candidate set $\mathcal{C}^m$ from $\mathcal{U}$
		\State Remove candidates violating time-window, attitude-maneuvering, energy, or storage constraints
		\If{$\mathcal{C}^m=\emptyset$}
		\State \textbf{break}
		\EndIf
		\State Compute the selection probability $P_m(z|u)$ for each $z\in\mathcal{C}^m$
		\State Select a candidate node $z=(i_z,s_z,k_z)$ by roulette-wheel selection
		\State Insert $z$ into the satellite-specific sequence $\Pi^m_{s_z}$
		\State Update the state of satellite $s_z$, including its latest task and resource states
		\State Update the unscheduled task set $\mathcal{U}\leftarrow\mathcal{U}\setminus\{i_z\}$
		\EndWhile
		\State Merge all satellite-specific sequences into $\Pi^m$
		\State \Return $\Pi^m$
	\end{algorithmic}
\end{algorithm}

The heuristic value combines benefit contribution, energy-efficiency contribution, and workload balance:
\begin{equation}
	\eta_m(u,z)
	=
	\chi_1G(u,z)+\chi_2E_m(u,z)+\chi_3B_m(z),
\end{equation}
where $G(u,z)$ denotes the benefit contribution of selecting candidate node $z$ after task $u$, $E_m(u,z)$ denotes the energy-efficiency contribution considering the additional observation and maneuvering energy, and $B_m(z)$ denotes the workload-balance contribution after inserting $z$. The coefficients $\chi_1$, $\chi_2$, and $\chi_3$ are nonnegative heuristic weights. The three terms are normalized before aggregation.

After all ants have constructed their schedules, each schedule is evaluated by the objective function in Section~\ref{sec02}. The pheromone matrix is then updated by evaporation and solution-quality-based deposition:
\begin{equation}
	\tau_{ij}
	\leftarrow
	\max\{(1-\rho)\tau_{ij},\tau_{\min}\},
	\quad
	\tau_{ij}
	\leftarrow
	\min\{\tau_{ij}+\Delta\tau_{ij},\tau_{\max}\},
\end{equation}
with the deposition increment
\begin{equation}
	\Delta\tau_{ij}
	=
	\sum_{m=1}^{N_A}
	\frac{F^m}{F_{\max}}
	I\bigl((i,j)\in\Pi^m\bigr),
\end{equation}
where $N_A$ is the number of ants, $F^m$ is the objective value of schedule $\Pi^m$, $F_{\max}$ is the current best objective, and $I(\cdot)$ is the indicator function.

\subsection{Markov Decision Process (MDP) Formulation}

The adaptive control of ACO parameters is formulated as an MDP
$\mathcal{M}=(\mathcal{X},\mathcal{A},\mathcal{P},R,\gamma)$,
where each decision step corresponds to one ACO iteration. After the $t$th iteration, the search state is extracted from the pheromone matrix and the current optimization results. The IQL policy then outputs a continuous action to adjust the pheromone factor $\alpha$, heuristic factor $\beta$, and evaporation rate $\rho$ for the next iteration.

The search state is defined as
\begin{equation}
    \mathbf{s}_t=
    \left[
    \bar{\tau}_t,\,
    \sigma_{\tau,t}^{2},\,
    f_t^{\mathrm{cur}},\,
    f_t^{\mathrm{gb}},\,
    r_t^{\mathrm{iter}}
    \right]^T,
\end{equation}
where $\bar{\tau}_t$ and $\sigma_{\tau,t}^{2}$ are the normalized mean and variance of the pheromone matrix, respectively. The variables $f_t^{\mathrm{cur}}$ and $f_t^{\mathrm{gb}}$ denote the best objective value of the current iteration and the global best objective value found so far, and $r_t^{\mathrm{iter}}$ is the normalized iteration ratio. These variables describe the pheromone distribution, current solution quality, historical best performance, and search progress.

The action is a continuous parameter-adjustment vector:
\begin{equation}
	\mathbf{a}_t=
	\left[
	\Delta\alpha_t,\,
	\Delta\beta_t,\,
	\Delta\rho_t
	\right]^{T}.
\end{equation}
After receiving the action, the ACO parameters are updated as
\begin{align}
	\alpha_{t+1}
	&=
	\operatorname{clip}(\alpha_t+\Delta\alpha_t,\alpha_{\min},\alpha_{\max}),\\
	\beta_{t+1}
	&=
	\operatorname{clip}(\beta_t+\Delta\beta_t,\beta_{\min},\beta_{\max}),\\
	\rho_{t+1}
	&=
	\operatorname{clip}(\rho_t+\Delta\rho_t,\rho_{\min},\rho_{\max}),
\end{align}
where $\operatorname{clip}(\cdot)$ restricts each parameter to its feasible range.

The reward encourages solution improvement while maintaining search diversity. Let $f_{t+1}^{\mathrm{cur}}$ be the best objective after applying the adjusted parameters. The improvement term is
\begin{equation}
	I_t=
	\max\left(0,\,
	f_{t+1}^{\mathrm{cur}}-f_t^{\mathrm{gb}}
	\right).
\end{equation}
The final reward is formulated as
\begin{equation}
	R_t=
	\operatorname{clip}
	\left(
	c_I I_t+\xi D_t+\kappa B_t,\,
	R_{\min},R_{\max}
	\right),
\end{equation}
where $D_t$ is a diversity-related term that measures the dispersion of the current search process, and $B_t$ is a bonus term activated when a new global-best solution is obtained. The coefficients $c_I$, $\xi$, and $\kappa$ balance immediate objective improvement, diversity preservation, and global progress. The clipping operation limits excessively large learning targets and improves the stability of offline training.

\subsection{Offline IQL Training}

The offline dataset is collected by running ACO with exploratory parameter adjustments on training scenarios. During data collection, ACO first performs one iteration to obtain the initial search state. At each subsequent decision step, an exploratory action is sampled to update $\alpha$, $\beta$, and $\rho$, and the next ACO iteration is executed with the updated parameters. The resulting reward, next state, and terminal indicator are recorded. The dataset is written as
\begin{equation}
	\mathcal{D}=\{(\mathbf{s}_t,\mathbf{a}_t,R_t,\mathbf{s}_{t+1},d_t)\}_{t=1}^{N_D},
\end{equation}
where $d_t$ is the terminal indicator and $N_D$ is the number of collected transitions. Algorithm~\ref{alg02} summarizes the transition collection process.

\begin{algorithm}[!t]
	\caption{Offline Transition Collection for IQL}
	\label{alg02}
	\begin{algorithmic}[1]
		\Require Training scenarios, maximum iteration number $T_{\max}$, and exploratory action strategy
		\Ensure Offline dataset $\mathcal{D}$
		
		\State Initialize $\mathcal{D}\leftarrow\emptyset$
		\For{each training scenario}
		\For{each training episode}
		\State Initialize an ACO scheduler
		\State Run one ACO iteration and extract the initial state $\mathbf{s}_0$
		\For{$t=0$ to $T_{\max}-2$}
		\State Extract the current state $\mathbf{s}_t$
		\State Sample an exploratory action $\mathbf{a}_t$
		\State Update $\alpha$, $\beta$, and $\rho$ using $\mathbf{a}_t$
		\State Run the next ACO iteration with the updated parameters
		\State Compute the reward $R_t$
		\State Extract the next state $\mathbf{s}_{t+1}$
		\State Determine the terminal indicator $d_t$
		\State Store $(\mathbf{s}_t,\mathbf{a}_t,R_t,\mathbf{s}_{t+1},d_t)$ in $\mathcal{D}$
		\EndFor
		\EndFor
		\EndFor
		\State \Return $\mathcal{D}$
	\end{algorithmic}
\end{algorithm}

IQL learns a value function $V_{\psi}(\mathbf{s})$, two Q-functions $Q_{\vartheta_1}(\mathbf{s},\mathbf{a})$ and $Q_{\vartheta_2}(\mathbf{s},\mathbf{a})$, and a policy function $\pi_{\phi}(\mathbf{s})$ from $\mathcal{D}$. The Q-learning target is $y_t=R_t+\gamma(1-d_t)V_{\bar{\psi}}(\mathbf{s}_{t+1})$, where $V_{\bar{\psi}}$ is the target value network. The Q loss is
\begin{equation}
	\mathcal{L}_{Q}(\vartheta_1,\vartheta_2)
	=
	\sum_{j=1}^{2}
	\mathbb{E}_{\mathcal{D}}
	\left[
	\left(
	Q_{\vartheta_j}(\mathbf{s}_t,\mathbf{a}_t)-y_t
	\right)^2
	\right].
\end{equation}

The value function uses expectile regression with $\hat{Q}=\min_j Q_{\vartheta_j}$:
\begin{equation}
\begin{split}
\mathcal{L}_{V}(\psi)
&=
\mathbb{E}_{\mathcal{D}}
\left[
L_{\tau_e}
\left(
\hat{Q}(\mathbf{s}_t,\mathbf{a}_t)-V_{\psi}(\mathbf{s}_t)
\right)
\right],\\
L_{\tau_e}(u)&=|\tau_e-\mathbb{I}(u<0)|u^2 .
\end{split}
\end{equation}

The policy is trained by advantage-weighted regression with $A(\mathbf{s}_t,\mathbf{a}_t)=\hat{Q}(\mathbf{s}_t,\mathbf{a}_t)-V_{\psi}(\mathbf{s}_t)$:
\begin{equation}
	\mathcal{L}_{\pi}(\phi)
	=
	\mathbb{E}_{\mathcal{D}}
	\left[
	\exp\left(\frac{A(\mathbf{s}_t,\mathbf{a}_t)}{\lambda}\right)
	\left\|
	\pi_{\phi}(\mathbf{s}_t)-\mathbf{a}_t
	\right\|_2^2
	\right].
\end{equation}

\subsection{Overall IQACO Procedure}

Algorithm~\ref{alg03} summarizes IQACO. In the offline stage, transitions are collected and used to train the IQL networks; the policy $\pi_{\phi}$ is exported. In the online stage, ants construct schedules using current $\alpha$ and $\beta$, the best schedule and pheromone matrix are updated with $\rho$, and $\pi_{\phi}$ outputs parameter adjustments $[\Delta\alpha,\Delta\beta,\Delta\rho]^T$ for the next iteration.

\begin{algorithm}[!t]
	\caption{IQL-Bootstrapped Ant Colony Optimization}
	\label{alg03}
	\begin{algorithmic}[1]
		\Require Training scenarios, test scenario, ACO settings, maximum iteration number $T_{\max}$, and number of ants $N_A$
		\Ensure Best schedule $\Pi^{\ast}$
		
		\State \textbf{Offline training stage}
		\State Collect transition dataset $\mathcal{D}$ using Algorithm~\ref{alg02}
		\State Train the IQL networks $V_{\psi}$, $Q_{\vartheta_1}$, $Q_{\vartheta_2}$, and $\pi_{\phi}$ using $\mathcal{D}$
		\State Export the trained policy $\pi_{\phi}$
		
		\Statex
		\State \textbf{Online scheduling stage}
		\State Initialize the ACO scheduler, pheromone matrix $\boldsymbol{\tau}$, parameters $\alpha$, $\beta$, $\rho$, and best schedule $\Pi^{\ast}\leftarrow\emptyset$
		\For{$t=0$ to $T_{\max}-1$}
		\For{$m=1$ to $N_A$}
		\State Construct a feasible schedule $\Pi^m$ using Algorithm~\ref{alg01}
		\EndFor
		\State Evaluate all schedules by the objective function
		\State Update the best schedule $\Pi^{\ast}$
		\State Update the pheromone matrix using $\rho$
		\If{$t<T_{\max}-1$}
		\State Extract the search state $\mathbf{s}_t$
		\State Obtain the action $\mathbf{a}_t=\pi_{\phi}(\mathbf{s}_t)$
		\State Update $\alpha$, $\beta$, and $\rho$ for the next iteration
		\EndIf
		\EndFor
		\State \Return $\Pi^{\ast}$
	\end{algorithmic}
\end{algorithm}

The policy input is five-dimensional, and policy inference is performed once per ACO iteration before updating the three ACO parameters.

\begin{table}[!t]
	\centering
	\caption{Configuration of the experimental scenarios}
	\label{tab02}
	\footnotesize
	\setlength{\tabcolsep}{4pt}
	\begin{tabular}{ccc}
		\toprule
		Scenario & Original targets & Satellites  \\
		\midrule
		Scene 01 & 100 & 3  \\
		Scene 02 & 120 & 3  \\
		Scene 03 & 100 & 4  \\
		Scene 04 & 120 & 4  \\
		Scene 05 & 140 & 4  \\
		Scene 06 & 160 & 4  \\
		Scene 07 & 140 & 5  \\
		Scene 08 & 160 & 5  \\
		Scene 09 & 180 & 5  \\
		Scene 10 & 200 & 5  \\
		Scene 11 & 180 & 6  \\
		Scene 12 & 200 & 6  \\
		Scene 13 & 220 & 6  \\
		Scene 14 & 240 & 6  \\
		\bottomrule
	\end{tabular}
\end{table}

\section{Simulation Experiments}\label{sec04}

All algorithms are implemented in C++20 (compiled with \texttt{-std=c++20 -O2} using MinGW-w64 GCC 13.2.0) and executed on a workstation with an Intel Core Ultra 9 285H and 64 GB RAM. The IQL module is implemented in Python with PyTorch; the trained policy is exported in ONNX format and loaded via the ONNX Runtime C++ API for CPU inference.

\subsection{Scenario Configuration}

Fourteen testing scenarios are constructed with 100--240 maritime moving targets and 3--6 satellites, as listed in Table~\ref{tab02}. Each original target has 2--4 observation requirements, which are expanded into independent scheduling tasks. The IQL policy is trained on separately generated scenarios with the same parameter ranges but different target distributions and satellite initial conditions, while the 14 scenarios in Table~\ref{tab02} are used only for testing.

Targets move within a representative East Asian domain
($\varphi\in[6^\circ,45^\circ]$, $\lambda\in[105^\circ,145^\circ]$)
over a 24-h planning horizon with a 1-s trajectory step. The target speed is sampled from $U(5,15)$ m/s, and the direction-persistence probability is sampled from $U(0.70,0.95)$. Each observation requirement has a duration of $d_i=60$ s and nominal benefit $p_i\in\{1,2,3\}$.

The satellite parameters are fixed across all scenarios. The orbital altitude is 400 km and the observation cone angle is $25^\circ$. The onboard energy capacity is $E_s^{\max}=500$ Wh, with imaging power $P_s^{\mathrm{img}}=750$ W and attitude-control power $P_s^{\mathrm{att}}=30$ W. The storage capacity is $M_s^{\max}=2000$ GB and the data-generation rate is $R_s^{\mathrm{data}}=4.0$ Gbps. Within the planning horizon, energy and storage are treated as fixed onboard budgets; energy recharge and data downlink are not modeled.
Cloud-affected availability is assigned according to target latitude: $c_{i,s,k}=0.60$ for $|\varphi|<10^\circ$, $0.70$ for $10^\circ\leq|\varphi|<25^\circ$, and $0.80$ for $25^\circ\leq|\varphi|<45^\circ$. The availability factor is used in both candidate-window screening and objective evaluation.

\subsection{Compared Algorithms and Parameter Settings}

IQACO is compared with GA, PSO, WOA, conventional ACO~\cite{33,34,35}, and the learning-enhanced RLGA~\cite{36}. For fairness, all algorithms use the same scheduling model, objective function, constraint-checking procedure, and evaluation budget. Unless otherwise stated, the objective weights are fixed at $(\eta_1,\eta_2,\eta_3)=(0.60,0.25,0.15)$, and the reported quantity is the weighted objective value $F$ defined in Section~\ref{sec02}. Each algorithm is independently executed 20 times with the same evaluation budget of $N_{\mathrm{FE}}=20000$. The parameter settings of the baseline algorithms follow commonly used configurations reported in the literature~\cite{33,34,35,36}, without additional problem-specific tuning.

For IQACO, the initial ACO parameters are $\alpha=1.0$, $\beta=2.5$, and $\rho=0.1$. The IQL policy outputs $[\Delta\alpha,\Delta\beta,\Delta\rho]^T$ with $\Delta\alpha\in[-0.2,0.2]$, $\Delta\beta\in[-0.4,0.4]$, and $\Delta\rho\in[-0.1,0.1]$, clipped to $\alpha\in[1.0,5.0]$, $\beta\in[1.0,5.0]$, and $\rho\in[0.1,0.5]$. IQL hyperparameters are listed in Table~\ref{tab03}.

\begin{table}[!t]
	\centering
	\caption{IQL training hyperparameters}
	\label{tab03}
	\footnotesize
	\setlength{\tabcolsep}{3pt}
	\begin{tabular}{p{0.22\columnwidth}p{0.47\columnwidth}p{0.18\columnwidth}}
		\toprule
		Parameter & Description & Value \\
		\midrule
		$\gamma$ & Discount factor & 0.99 \\
		$\tau_e$ & Expectile parameter & 0.6 \\
		$\lambda$ & Inverse temperature & 5.0 \\
		State dimension & Input features & 5 \\
		Action dimension & Parameter-adjustment action & 3 \\
		Epochs & Training epochs & 200 \\
		Learning rate & Adam optimizer & $3\times10^{-4}$ \\
		\bottomrule
	\end{tabular}
\end{table}

\subsection{Results and Discussion}

\subsubsection{Convergence Analysis}\label{subsec:conv}

\begin{figure}[H]
	\centering
	\includegraphics[width=0.6\linewidth]{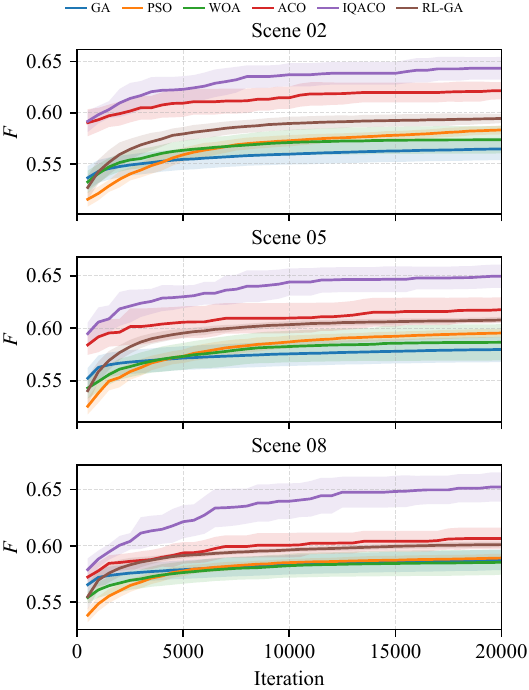}
	\caption{Convergence comparison on Scenes 02, 05, and 08. Each curve is the mean best-so-far weighted objective value $F$ over 20 independent runs, and the shaded band denotes $\pm1$ standard deviation.}
	\label{fig03}
\end{figure}

\begin{figure}[H]
	\centering
	\includegraphics[width=0.6\linewidth]{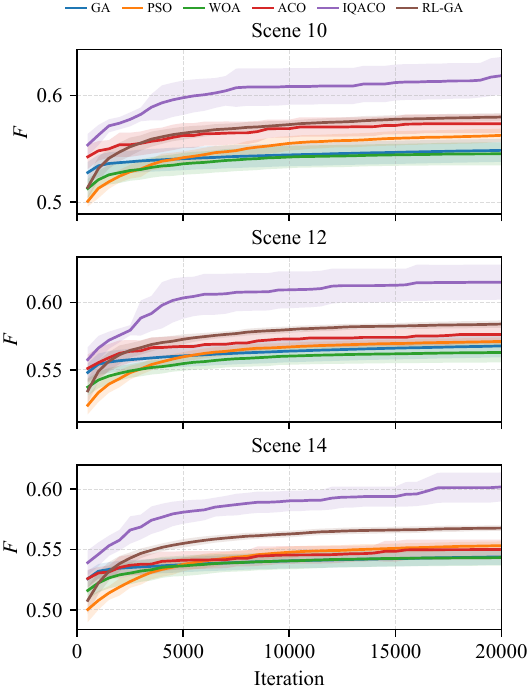}
	\caption{Convergence comparison on Scenes 10, 12, and 14. Each curve is the mean best-so-far weighted objective value $F$ over 20 independent runs, and the shaded band denotes $\pm1$ standard deviation.}
	\label{fig04}
\end{figure}

\begin{figure}[H]
	\centering
	\includegraphics[width=0.6\linewidth]{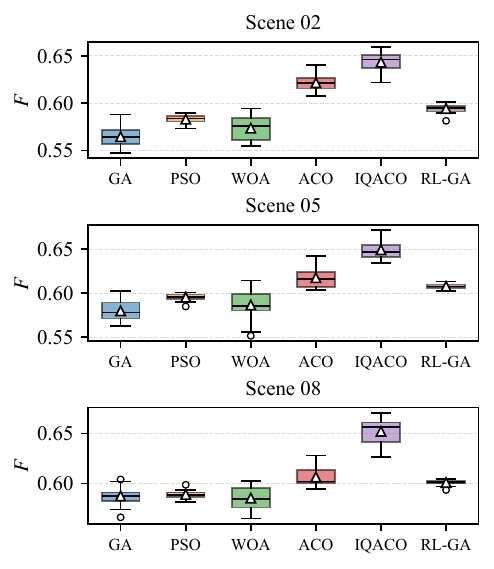}
\caption{Distribution of the final weighted objective value $F$ on Scenes 02, 05, and 08. Each box summarizes 20 independent runs; the box spans the interquartile range, the horizontal line is the median, the triangle is the mean, the whiskers extend to the most extreme values within 1.5 times the interquartile range, and open circles mark outliers.}
	\label{fig05}
\end{figure}

\begin{figure}[H]
	\centering
	\includegraphics[width=0.6\linewidth]{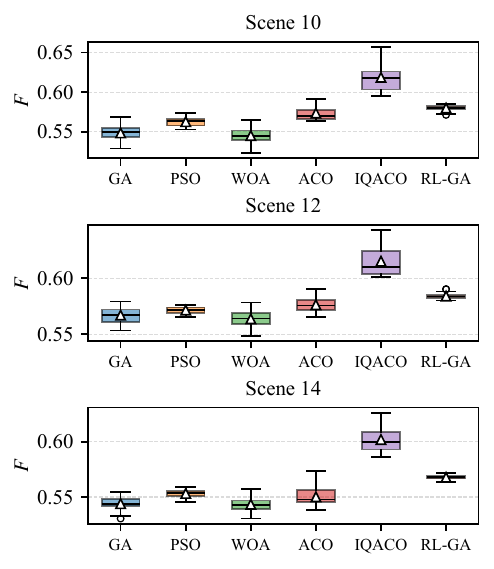}
\caption{Distribution of the final weighted objective value $F$ on Scenes 10, 12, and 14. Each box summarizes 20 independent runs; the box spans the interquartile range, the horizontal line is the median, the triangle is the mean, the whiskers extend to the most extreme values within 1.5 times the interquartile range, and open circles mark outliers.}
	\label{fig06}
\end{figure}

The convergence behavior of the six algorithms is examined on six representative scenarios spanning moderate and large problem scales. Each algorithm is independently executed 20 times per scenario, and the best-so-far objective value is recorded. Figs.~\ref{fig03} and~\ref{fig04} report the mean convergence curves with one-standard-deviation bands, whereas Figs.~\ref{fig05} and~\ref{fig06} show the corresponding final-objective distributions.

\begin{table}[!t]
	\centering
	\caption{Statistical Summary of the Final Weighted Objective Value $F$ Over 20 Independent Runs}
	\label{tab04}
	\scriptsize
	\setlength{\tabcolsep}{3.6pt}
	\renewcommand{\arraystretch}{1.10}
	\begin{threeparttable}
		\begin{tabular}{@{}c@{\hskip 2pt}c@{\hskip 2pt}c@{\hskip 2pt}c@{\hskip 2pt}c@{\hskip 2pt}c@{\hskip 2pt}c@{\hskip 2pt}c@{}}
			\toprule
				\textbf{Scene} & \textbf{GA} & \textbf{PSO} & \textbf{WOA} & \textbf{ACO} & \textbf{RLGA} & \textbf{IQACO} & \textbf{Gain} \\
			\midrule
				\textbf{01} & $0.5857\pm0.0128$ & $0.6003\pm0.0044$ & $0.5867\pm0.0154$ & $0.6348\pm0.0100$ & $0.6083\pm0.0028$ & $\mathbf{0.6564\pm0.0167}^{\ddagger}$ & $3.41\%$ \\
				\textbf{02} & $0.5647\pm0.0109$ & $0.5832\pm0.0044$ & $0.5737\pm0.0134$ & $0.6216\pm0.0084$ & $0.5944\pm0.0046$ & $\mathbf{0.6434\pm0.0110}^{\ddagger}$ & $3.51\%$ \\
				\textbf{03} & $0.6155\pm0.0122$ & $0.6159\pm0.0033$ & $0.6109\pm0.0108$ & $0.6481\pm0.0105$ & $0.6229\pm0.0038$ & $\mathbf{0.6666\pm0.0149}^{\ddagger}$ & $2.86\%$ \\
				\textbf{04} & $0.5900\pm0.0110$ & $0.6028\pm0.0034$ & $0.5884\pm0.0145$ & $0.6293\pm0.0127$ & $0.6102\pm0.0044$ & $\mathbf{0.6518\pm0.0152}^{\ddagger}$ & $3.57\%$ \\
				\textbf{05} & $0.5798\pm0.0120$ & $0.5954\pm0.0040$ & $0.5866\pm0.0172$ & $0.6176\pm0.0119$ & $0.6078\pm0.0029$ & $\mathbf{0.6494\pm0.0111}^{\ddagger}$ & $5.14\%$ \\
				\textbf{06} & $0.5588\pm0.0104$ & $0.5774\pm0.0061$ & $0.5565\pm0.0118$ & $0.5985\pm0.0121$ & $0.5949\pm0.0041$ & $\mathbf{0.6343\pm0.0130}^{\ddagger}$ & $5.99\%$ \\
				\textbf{07} & $0.6038\pm0.0085$ & $0.6052\pm0.0025$ & $0.6026\pm0.0065$ & $0.6263\pm0.0093$ & $0.6128\pm0.0029$ & $\mathbf{0.6719\pm0.0109}^{\ddagger}$ & $7.28\%$ \\
				\textbf{08} & $0.5876\pm0.0091$ & $0.5891\pm0.0038$ & $0.5854\pm0.0111$ & $0.6066\pm0.0096$ & $0.6011\pm0.0027$ & $\mathbf{0.6523\pm0.0130}^{\ddagger}$ & $7.54\%$ \\
				\textbf{09} & $0.5616\pm0.0094$ & $0.5679\pm0.0031$ & $0.5551\pm0.0080$ & $0.5793\pm0.0085$ & $0.5805\pm0.0037$ & $\mathbf{0.6254\pm0.0147}^{\ddagger}$ & $7.97\%$ \\
				\textbf{10} & $0.5485\pm0.0108$ & $0.5626\pm0.0062$ & $0.5452\pm0.0107$ & $0.5735\pm0.0087$ & $0.5797\pm0.0037$ & $\mathbf{0.6186\pm0.0175}^{\ddagger}$ & $7.86\%$ \\
				\textbf{11} & $0.5731\pm0.0084$ & $0.5717\pm0.0033$ & $0.5698\pm0.0067$ & $0.5872\pm0.0098$ & $0.5823\pm0.0019$ & $\mathbf{0.6381\pm0.0132}^{\ddagger}$ & $8.66\%$ \\
				\textbf{12} & $0.5669\pm0.0075$ & $0.5714\pm0.0033$ & $0.5637\pm0.0078$ & $0.5762\pm0.0068$ & $0.5840\pm0.0027$ & $\mathbf{0.6151\pm0.0129}^{\ddagger}$ & $6.75\%$ \\
				\textbf{13} & $0.5580\pm0.0094$ & $0.5654\pm0.0037$ & $0.5570\pm0.0080$ & $0.5697\pm0.0081$ & $0.5777\pm0.0026$ & $\mathbf{0.6158\pm0.0163}^{\ddagger}$ & $8.10\%$ \\
				\textbf{14} & $0.5440\pm0.0065$ & $0.5532\pm0.0035$ & $0.5434\pm0.0067$ & $0.5502\pm0.0082$ & $0.5678\pm0.0022$ & $\mathbf{0.6019\pm0.0120}^{\ddagger}$ & $9.41\%$ \\
			\bottomrule
		\end{tabular}
		\begin{tablenotes}
    \footnotesize
    \item[$\ddagger$] indicates that IQACO is significantly better than all five baseline algorithms at $p<0.05$ according to the Wilcoxon signed-rank test.
\end{tablenotes}
	\end{threeparttable}
\end{table}

Figs.~\ref{fig03} and~\ref{fig05} present the convergence behavior and final-objective distributions for the moderate-scale cases (Scenes 02, 05, and 08). IQACO achieves the highest mean objective values in all three scenes, followed by conventional ACO and RLGA. Compared with conventional ACO, IQACO also reaches higher best-so-far values and higher final-objective distributions over repeated runs. RLGA obtains higher mean objective values than GA, PSO, and WOA in these three cases.

Fig.~\ref{fig04} shows the convergence results for the larger cases (Scenes 10, 12, and 14). IQACO reaches higher best-so-far objective values than the five baseline algorithms in all three scenes. The final-objective distributions in Fig.~\ref{fig06} show the same ordering. RLGA also performs better than conventional ACO in these cases, while its mean objective values remain below those of IQACO.

Table~\ref{tab04} reports the statistical results for all 14 scenarios. IQACO obtains the highest mean weighted objective value in every scenario, and the differences from the five baseline algorithms are significant at $p<0.05$ according to the Wilcoxon signed-rank test. 

The Gain column of Table~\ref{tab04} denotes the relative improvement of IQACO over conventional ACO:
\begin{equation}
	\mathrm{Gain}
	=
	\frac{\bar{F}_{\mathrm{IQACO}}-\bar{F}_{\mathrm{ACO}}}
	{\bar{F}_{\mathrm{ACO}}}
	\times100\%,
\end{equation}
where $\bar{F}$ is the mean final weighted objective value over 20 independent runs. All Gain values are calculated from the unrounded means. The improvement ranges from $2.86\%$ in Scene~03 to $9.41\%$ in Scene~14.

\subsubsection{Weight Sensitivity Analysis}\label{subsec:weight}

\begin{figure}[H]
	\centering
	\includegraphics[width=0.40\textwidth]{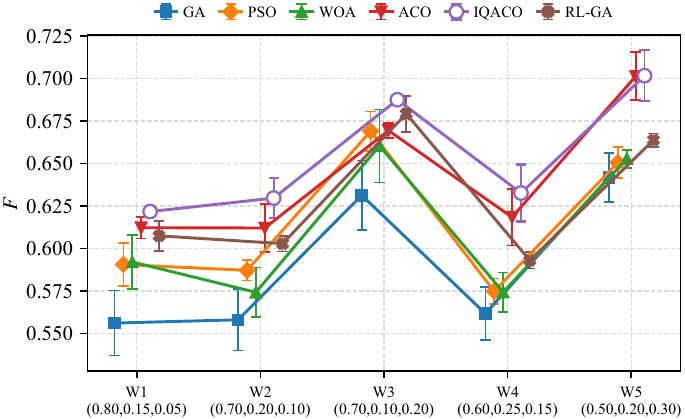}
	\includegraphics[width=0.40\textwidth]{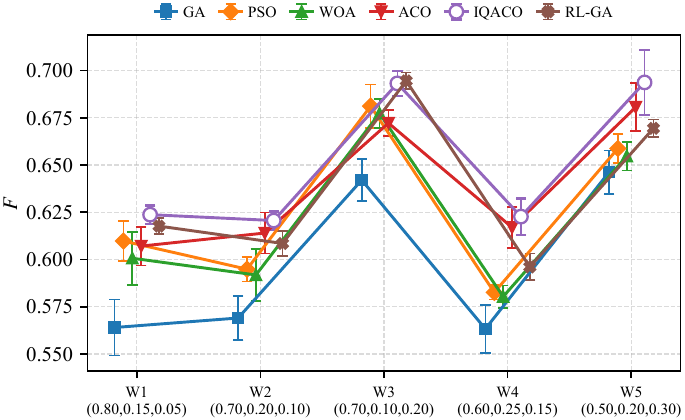}
	\includegraphics[width=0.40\textwidth]{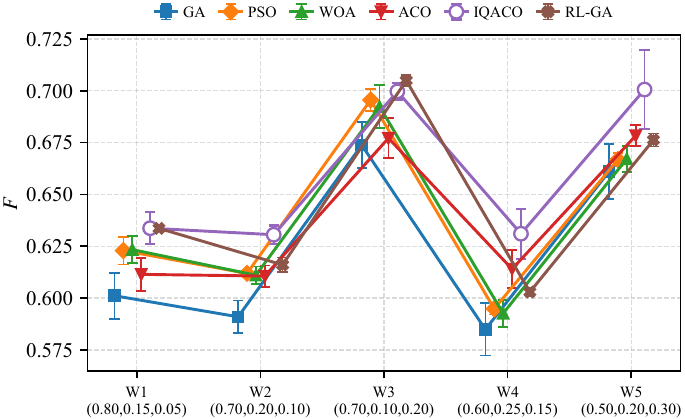}
	\includegraphics[width=0.40\textwidth]{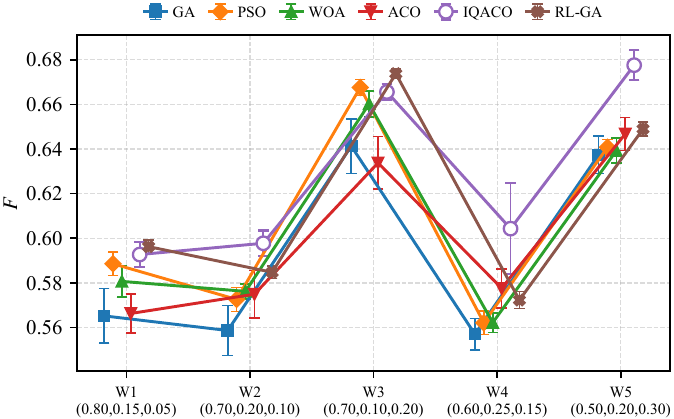}
\caption{Weight sensitivity on Scenes 01, 04, 07, and 11 under the five asymmetric weight configurations W1--W5. Each point denotes the mean weighted objective value $F$ over 10 independent runs, and the error bar represents $\pm1$ standard deviation. The weights are ordered as $(\eta_1,\eta_2,\eta_3)$.}
	\label{fig07}
\end{figure}

Weight sensitivity is evaluated on Scenes 01, 04, 07, and 11 using five asymmetric weight configurations: W1 $=(0.80,0.15,0.05)$, W2 $=(0.70,0.20,0.10)$, W3 $=(0.70,0.10,0.20)$, W4 $=(0.60,0.25,0.15)$, and W5 $=(0.50,0.20,0.30)$, where the weights are ordered as $(\eta_1,\eta_2,\eta_3)$. W4 is the setting used in the main experiments. W2 and W3 keep $\eta_1=0.70$ unchanged while exchanging the relative weights of energy efficiency and workload balance, allowing the effect of these two preferences to be compared directly. Fig.~\ref{fig07} reports the mean and standard deviation over 10 independent runs.

IQACO obtains the highest mean objective value in 16 of the 20 scene--configuration combinations, while RLGA gives the highest mean value in the remaining four. Relative to conventional ACO, the improvement of IQACO ranges from $0.05\%$ to $5.0\%$ across the tested configurations. The W2--W3 comparison shows that changing the relative preference between energy efficiency and workload balance affects the performance margin differently across the four scenes, with no consistent trend across problem instances.

\subsubsection{IQL Training Convergence Analysis}

\begin{figure}[H]
	\centering
	\includegraphics[width=0.98\linewidth]{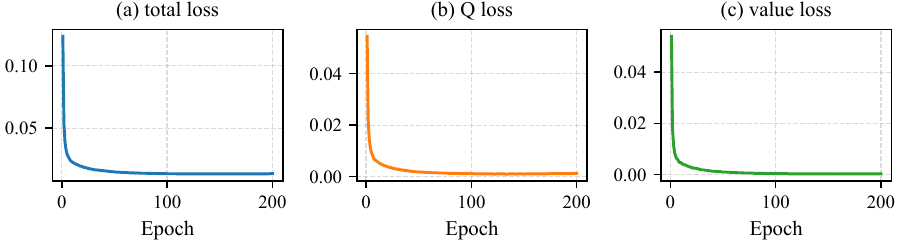}
	\caption{Training loss terms of the IQL module over 200 epochs: (a) total loss; (b) Q loss; (c) value loss.}
	\label{fig08}
\end{figure}

\begin{figure}[H]
	\centering
	\includegraphics[width=0.98\linewidth]{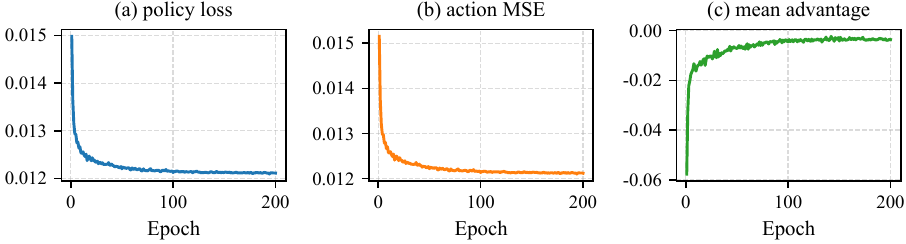}
	\caption{Policy and advantage indicators of the IQL module over 200 epochs: (a) policy loss; (b) action MSE; (c) mean advantage $A(\mathbf{s},\mathbf{a})=\hat{Q}(\mathbf{s},\mathbf{a})-V(\mathbf{s})$.}
	\label{fig09}
\end{figure}

\begin{figure}[H]
	\centering
	\includegraphics[width=0.98\linewidth]{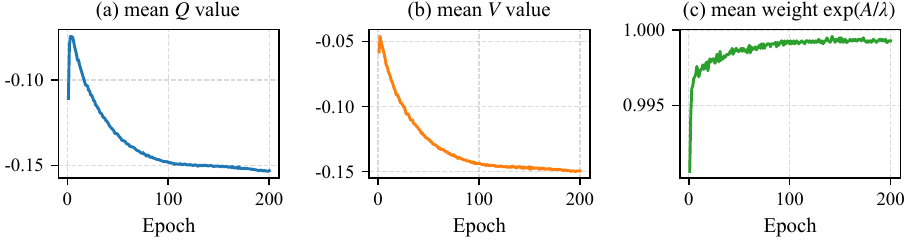}
	\caption{Value estimation and weight statistics of the IQL module over 200 epochs: (a) mean $Q$ value; (b) mean $V$ value; (c) mean advantage-weighted regression weight $\exp(A/\lambda)$.}
	\label{fig10}
\end{figure}

To further examine the training behavior of the IQL module, the main loss functions and learning statistics are recorded during offline training. Figs.~\ref{fig08}--\ref{fig10} show the variations of the total loss, Q loss, value loss, policy loss, action mean-square error, mean Q value, mean V value, mean advantage, and mean action weight with respect to the training epoch.

The training curves show that the IQL module remains stable during offline learning. In the early training stage, the total loss and the main component losses decrease rapidly, indicating that the networks learn useful state--action value information from the offline transition samples. As the number of epochs increases, these losses gradually enter a bounded fluctuation range, and no evident divergence is observed. The decreases in policy loss and action mean-square error indicate that the policy network gradually fits parameter-adjustment actions with higher estimated advantages.

The mean Q value and mean V value also become more stable in the later training stage, suggesting that the value estimates converge to relatively consistent levels. The mean advantage and mean action weight remain within reasonable ranges, which indicates that the advantage-weighted regression does not produce excessively large sample weights. Overall, the training curves show that the IQL losses and value estimates remain bounded over the 200 training epochs, and the learned policy is subsequently used for parameter adjustment in the test scenarios.

\section{Conclusion}\label{sec05}

This article addressed multi-satellite maritime moving-target observation scheduling using an IQL-guided adaptive ACO framework. The IQL policy adjusts the three ACO parameters according to a five-dimensional search state, while feasible schedules are constructed by the ACO decoder. In the main comparison over 14 scenarios, IQACO achieves the highest mean weighted objective value in every scenario under the main objective-weight setting, with an improvement of $2.86\%$--$9.41\%$ over conventional ACO. The asymmetric weight experiments further examine the effect of different objective preferences on the relative performance of the algorithms. 
The current study considers a single planning horizon, and the residual energy and storage at the end of the horizon are not carried into subsequent horizons. In continuous multi-period scheduling, these residual resources may affect feasible decisions and scheduling performance in later periods. Future work will therefore extend the formulation to a rolling-horizon framework in which residual energy and storage, together with downlink opportunities, are updated between planning periods.

\section*{Acknowledgements}

This work was supported by the National Natural Science Foundation of China
(Nos. 62373117, 62403158, 62573150, 62503128 and 72501042), the Fundamental Research Funds for the Central Universities (No. 3072025GH0401) and the Hainan Provincial Natural Science Foundation of China (No. 626QN0688).

\section*{Declaration of competing interest}
The authors declare that they have no known competing financial interests or personal relationships that could have appeared to influence the work reported in this paper.

\section*{Data availability}
The data and code supporting the findings of this study are available from the corresponding author upon reasonable request.

\end{document}